\documentclass[unnumsec,webpdf,contemporary,large, numbers]{oup-authoring-template}%

\graphicspath{{Fig/}}

\theoremstyle{thmstyleone}%
\theoremstyle{thmstyletwo}%
\theoremstyle{thmstylethree}%

\usepackage[most]{tcolorbox}
\usepackage{tabularx}
\usepackage{booktabs}
\usepackage{makecell}
\usepackage{graphicx} 
\usepackage{array}
\usepackage{amssymb}
\usepackage{pifont}

\begin{document}

\journaltitle{Submitted to xxx}
\DOI{}
\copyrightyear{2026}
\pubyear{2026}
\access{}
\appnotes{Paper}

\firstpage{1}


\title[]{Benchmarking AI scientists for omics data--driven biological discovery}

\author[1]{Erpai Luo}
\author[1]{Jinmeng Jia}
\author[1]{Yifan Xiong}
\author[2]{Xiangyu Li}
\author[3]{Xiaobo Guo}
\author[3]{Baoqi Yu}
\author[1]{Minsheng Hao}
\author[1]{Lei Wei}
\author[1,4,$\ast$]{Xuegong Zhang}

\authormark{Luo et al.}

\address[1]{\orgdiv{MOE Key Laboratory of Bioinformatics and Bioinformatics Division of BNRIST, Department of Automation}, \orgname{Tsinghua University}, \orgaddress{\postcode{100084}, \state{Beijing}, \country{China}}}
\address[2]{\orgdiv{School of Software Engineering}, \orgname{Beijing Jiaotong University}, \orgaddress{\postcode{100044}, \state{Beijing}, \country{China}}}
\address[3]{\orgdiv{Department of Physiology and Pathophysiology, School of Basic Medical Sciences},\orgname{Capital Medical University}, \orgaddress{\postcode{100069},\state{Beijing}, \country{China}}}
\address[4]{\orgdiv{Center for Synthetic and Systems Biology, School of Life Sciences and School of Medicine}, \orgname{Tsinghua University}, \orgaddress{\postcode{100084}, \state{Beijing}, \country{China}}}

\corresp[$\ast$]{Corresponding author}



\abstract{
\textbf{Motivation:} Recent advances in large language models have enabled the emergence of AI scientists that aim to autonomously analyze biological data and assist scientific discovery. Despite rapid progress, it remains unclear to what extent these systems can extract meaningful biological insights from real experimental data. Existing benchmarks either evaluate reasoning in the absence of data or focus on predefined analytical outputs, failing to reflect realistic, data-driven biological research.\\
\textbf{Results:} Here, we introduce BAISBench (Biological AI Scientist Benchmark), a benchmark for evaluating AI scientists on real single-cell transcriptomic datasets. BAISBench comprises two tasks: cell type annotation across 15 expert-labeled datasets, and scientific discovery through 193 multiple-choice questions derived from biological conclusions reported in 41 published single-cell studies. 
We evaluated several representative AI scientists using BAISBench and, to provide a human performance baseline, invited six graduate-level bioinformaticians to  collectively complete the same tasks.
The results show that while current AI scientists fall short of fully autonomous biological discovery, 
they already demonstrate substantial potential in supporting data-driven biological research.
These results position BAISBench as a practical benchmark for characterizing the current capabilities and limitations of AI scientists in biological research.
We expect BAISBench to serve as a practical evaluation framework for guiding the development of more capable AI scientists and for helping biologists identify AI systems that can effectively support real-world research workflows.\\
\textbf{Availability:} \href{https://github.com/EperLuo/BAISBench}{https://github.com/EperLuo/BAISBench}, \href{https://huggingface.co/datasets/EperLuo/BaisBench}{https://huggingface.co/datasets/EperLuo/BaisBench}.\\
\textbf{Contact:} \href{name@email.com}{zhangxg@tsinghua.edu.cn}\\
}


\maketitle

\section{Introduction}

Biological research increasingly relies on large-scale, data-driven analyses to elucidate cellular states, molecular mechanisms, and disease processes. Advances in high-throughput technologies have enabled the routine generation of complex datasets; however, transforming these data into biologically meaningful insight remains a labor-intensive and expertise-dependent process. From quality control and normalization to downstream interpretation and hypothesis formulation, successful analysis requires not only technical proficiency but also substantial domain expertise and biological insights. As datasets continue to grow in size and complexity, the analytical burden placed on individual researchers has emerged as a major bottleneck in modern biological research.

This challenge is especially pronounced in omics-driven studies such as transcriptomics, proteomics, and metabolomics. Among these, single-cell transcriptomics has emerged as a cornerstone technology, enabling molecular profiling at single-cell resolution and providing unprecedented access to cellular heterogeneity, rare populations, and dynamic biological processes across tissues and disease states \citep{svensson2018exponential, zheng2017massively}. 
A typical single-cell RNA sequencing (scRNA-seq) analysis involves multiple stages, including data preprocessing, dimensionality reduction, clustering, and cell type annotation, followed by downstream analyses such as differential gene expression, trajectory inference, and cell--cell communication analysis. While many of these steps are supported by mature computational pipelines, researchers must still carefully select appropriate tools and parameters and integrate analytical results with prior biological knowledge to derive biologically coherent conclusions \citep{luecken2019current, heumos2023best}. All these  processes remain difficult to automate and continue to rely heavily on expert judgment.

Recent advances in large language models (LLMs) have renewed interest in intelligent systems that aim to assist or automate components of the biological research workflow. Enabled by tool use, code generation, and interaction with external knowledge sources, a new class of systems, often referred to as AI scientists, has been proposed to autonomously analyze experimental data, reason about biological phenomena, and generate candidate scientific conclusions\citep{gottweis2025towards, lu2024ai, yamada2025ai, su2024two}. 
Early studies suggest that such systems can support data analysis and hypothesis generation, raising the possibility that parts of the biological discovery process may be, at least partially, automated \citep{zhang2024comprehensive, reddy2025towards}.

Motivated by this vision, a growing number of AI scientist systems have been developed for biological research.
For example,
AutoBA \citep{zhou2024ai} utilizes an LLM agent to construct automated pipelines for preprocessing and analyzing single-cell data. 
scChat \citep{lu2024scchat} combines a multi-agent architecture with retrieval-augmented generation to enable autonomous exploration of single-cell datasets. 
BioChatter \citep{lobentanzer2025platform} provides a generic backend framework for integrating conversational AI with biomedical tools. Biomni \citep{huang2025biomni} introduces a general-purpose AI scientist framework that integrates literature retrieval, tool use, and reasoning to support autonomous biomedical workflows. Pantheon \citep{pantheon} enables structured scientific reasoning and iterative experiment planning through coordinated LLM agents. STELLA \citep{jin2025stella} combines LLMs with tool-assisted data analysis to facilitate data-driven scientific discovery and hypothesis validation. 
Collectively, these efforts highlight rapid progress in system design and integration, but they also raise a critical question: what biological capabilities do current AI scientists actually possess?

Despite growing enthusiasm, the rigorous evaluation of AI scientists in realistic biological research settings remains an open challenge. 
Scientific discovery is inherently open-ended, and its quality is difficult to quantify using conventional metrics. 
Existing evaluation strategies often rely on expert review of AI-generated conclusions \citep{gottweis2025towards, schmidgall2025agent} or on LLM-based judges that approximate expert assessment \citep{baek2024researchagent}. While informative, these approaches are subjective, difficult to scale, and provide limited insight into whether an AI system can genuinely extract biologically meaningful knowledge from raw experimental data.

Several benchmarks have attempted to address this gap by evaluating AI agents on structured scientific tasks or question-answering settings \citep{qi2024large, laurent2024lab}. 
However, these benchmarks are primarily knowledge-driven and do not reflect the data-centric nature of biological research, where new insights emerge from the analysis of newly generated omics datasets rather than from static background information alone. 
The BLADE benchmark \citep{gu2024blade} partially addresses this limitation by incorporating data-driven questions, but focuses largely on domains outside molecular and cellular biology. 
BixBench \citep{mitchener2025bixbench} further grounds evaluation in real omics datasets across diverse biological scenarios, yet it mainly assesses computational outputs, without directly probing whether AI systems can produce interpretable, insight-rich biological conclusions. 
These limitations underscore the need for evaluation frameworks that align more closely with the core objectives of biological research: deriving biologically meaningful interpretations and discoveries from complex experimental data.

\begin{figure*}[!t]
  \centering
  \includegraphics[width=0.9\textwidth]{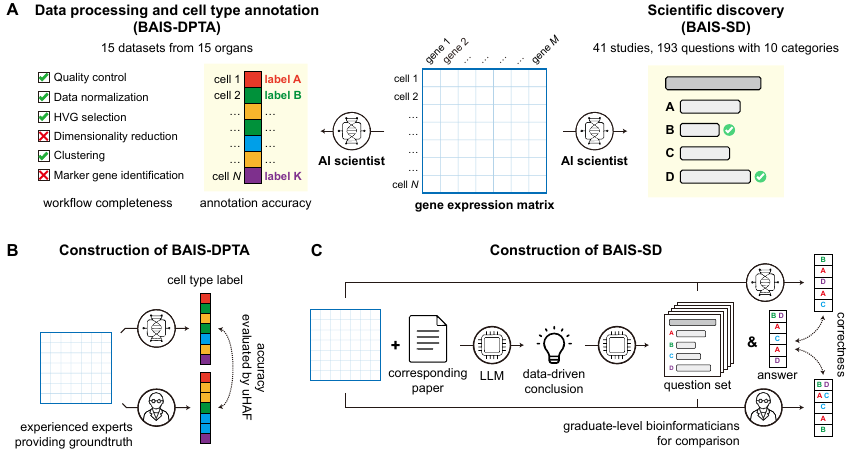}
  \caption{
(A) Overview of BAISBench.
(B) Construction of BAIS-DPTA: multi-organ single-cell transcriptomic datasets were curated and expert-annotated to support hierarchical cell type evaluation using uHAF.
(C) Construction of BAIS-SD: published single-cell studies and corresponding datasets were curated to derive data-driven multiple-choice questions, which were completed by both AI scientists and graduate-level bioinformaticians.}
  \label{main_fig}
\end{figure*}

Here, we introduce BAISBench (Biological AI Scientist Benchmark), a benchmark designed to evaluate AI scientists in realistic, data-driven biological discovery settings. BAISBench is built upon single-cell transcriptomic datasets, a representative and widely adopted modality in contemporary molecular and cellular biology. BAISBench consists of two complementary tasks (Figure~\ref{main_fig}A):

\begin{itemize}
  \item \textbf{Data Processing and cell Type Annotation (BAIS-DPTA)}: This task evaluates fundamental analytical capabilities using 15 expert-labeled single-cell datasets and introduces a hierarchical, cell type--aware metric that captures both annotation accuracy and biological granularity.
  \item \textbf{Scientific Discovery (BAIS-SD)}: This task comprises 193 multiple-choice questions derived from biological conclusions reported in 41 published single-cell studies, requiring AI scientists to analyze corresponding datasets and identify conclusions consistent with established findings.
\end{itemize}

Using BAISBench, we systematically evaluate several state-of-the-art AI scientists designed for single-cell analysis. 
To provide human performance baselines, we additionally involved six graduate-level bioinformaticians: one completed the BAIS-DPTA task using an automated cell annotation tool, and five collectively completed the BAIS-SD task.
Our results show that while current AI scientists remain less effective than human experts in tasks requiring deep biological judgment, they are already capable of reliably executing standard preprocessing and analytical workflows. Moreover, in the scientific discovery task, the best-performing AI scientists achieve performance comparable to that of graduate-level human researchers, with overall accuracy largely determined by the underlying base LLM models.

Together, these findings position BAISBench as a biologically grounded evaluation framework that characterizes both the capabilities and the limitations of current AI scientists in data-driven biological discovery. By providing a realistic and interpretable benchmark, BAISBench offers a principled basis for assessing progress in AI-assisted biology and for guiding the development and selection of AI systems that can meaningfully support real-world research workflows.

\section{Methods}

To evaluate AI scientists in realistic, data-driven biological research settings, we developed BAISBench, a benchmark composed of two complementary tasks that probe distinct but interrelated capabilities required for biological discovery (Figure \ref{main_fig}A).
The first task, Data Processing and Cell Type Annotation (BAIS-DPTA), assesses whether an AI scientist can reliably execute standard single-cell analytical workflows and apply biological knowledge to identify cell identities from gene expression profiles, without relying on external automated annotation tools such as CellTypist.
The second task, Scientific Discovery (BAIS-SD), evaluates higher-level biological reasoning by requiring AI scientists to analyze real single-cell datasets and answer questions derived from published biological discoveries.
Together, these two tasks are designed to decompose biological research into foundational analytical competence and integrative discovery-oriented reasoning, allowing systematic assessment of current AI scientists. The design and construction of both tasks are described in detail below.

\subsection{The data processing and cell type annotation (BAIS-DPTA) task}

The BAIS-DPTA task targets the most fundamental layer of single-cell data analysis, encompassing both computational execution and biologically informed interpretation. While many preprocessing steps can be standardized, accurate cell type annotation remains highly dependent on domain-specific biological knowledge and contextual judgment, making this task well suited for benchmarking core analytical capabilities of AI scientists.

As illustrated in Figure~\ref{main_fig}B, we curated 15 scRNA-seq datasets from recent publications, each representing a distinct organ or tissue. All datasets were selected to cover a broad range of tissue types, cell numbers (2,312–58,706), and annotation granularities (4–42 cell types). Detailed information on the source studies and datasets is provided in Table~\ref{task1data}. All datasets are provided in standardized \texttt{h5ad} format with raw gene expression counts.

To establish high-quality reference annotations, a team of experienced bioinformaticians manually annotated cell types in each dataset using the hierarchical taxonomy defined by the Unified Hierarchical cell Annotation Framework (uHAF) \citep{bian2025uhaf}. uHAF provides organ-specific hierarchical cell type trees (uHAF-Ts) constructed from anatomical literature, single-cell studies, and the Cell Ontology, covering 50 organs. This hierarchical structure enables cell type annotations to be compared at multiple levels of biological granularity and supports systematic evaluation beyond exact label matching.

\begin{figure*}[t]
  \centering
  \includegraphics[width=0.9\textwidth]{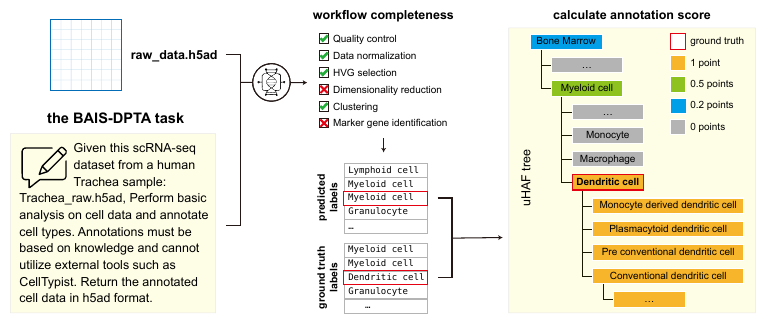}
  \caption{Pipeline of the BAIS-DPTA task. 
  The AI scientist is provided with a single-cell gene expression dataset from a specific organ and is required to perform cell type annotation using its own chosen method. The predicted annotations are then evaluated using a hierarchical scoring metric based on the uHAF cell type tree, which quantifies performance according to the granularity and correctness of the predictions.}
  \label{task1}
\end{figure*}

In BAIS-DPTA, each AI scientist is provided with a single dataset and is free to adopt any analysis strategy for preprocessing and cell type annotation (Figure~\ref{task1}). After completing the analysis, we first assess whether the AI scientist executed a complete and biologically reasonable preprocessing workflow. Specifically, we examine the presence of six essential steps commonly required in single-cell analysis: quality control, data normalization, highly variable gene (HVG) selection, dimensionality reduction, clustering, and marker gene identification.

We then evaluate cell type annotation performance  using a uHAF-based hierarchical score $S_{\rm CTA}$ that assigns full, partial, or weak credit to exact (or finer), parent, and grandparent (or equivalent higher-level ancestor)-level matches, respectively:

\begin{equation}
S_{\rm CTA} = \frac{n_{\rm exact} + 0.5 n_{\rm parent} + 0.2 n_{\rm grand}}{n_{\rm cell}}
\end{equation}
where $n_{\rm cell}$ is the total number of cells. This metric explicitly rewards biologically meaningful partial correctness and reflects the hierarchical nature of cell type definitions.

\subsection{The scientific discovery  (BAIS-SD) task}

For the BAIS-SD task, we constructed a set of multiple-choice questions derived from the biological discoveries reported in recently published single-cell research papers. This task is designed to evaluate the ability of AI scientists to generate biologically meaningful insights in realistic research scenarios. 
It requires not only effective data processing and analysis, but also appropriate tool usage, substantial domain-specific knowledge, and strong reasoning capabilities.

We curated 41 recently published single-cell studies together with their corresponding datasets from the CellxGene platform \citep{czi2025cz}. These datasets span a wide range of biological contexts and contain between 5,600 and 122,000 cells, along with metadata necessary for downstream analyses. A complete list of the studies and datasets is provided in Table~\ref{task2data}.

From each study, we constructed multiple-choice questions grounded in key biological findings reported in the original publication (Figure~\ref{main_fig}C). To ensure consistency and scalability, we used an LLM-assisted procedure to extract candidate discoveries and convert them into questions. Specifically, we provided each paper to GPT-4o and instructed it to:
\begin{enumerate}
    \item Give a quick and short summary of the research background in the first person, and extract the basic information about the sequencing data;
    \item Identify which  conclusions/discoveries in the article are derived directly from the single-cell transcriptomic data measured by the authors, and list them all;
    \item Identify which conclusions/discoveries in the article are based on a combination of data measured by the author and external knowledge, and list them all;
\item Select five appropriate conclusions/discoveries from the above and formulate them into multiple-choice questions.
\end{enumerate}

The full prompt is provided in Supplementary Materials.

The final BAIS-SD benchmark contains 193 questions, of which  approximately 20\% are multiple-choice. As shown in Figure~\ref{task2}B, questions span diverse analytical themes, including key gene analysis, cellular heterogeneity, functional interpretation, and developmental or disease-related reasoning. By grounding questions in published discoveries, BAIS-SD emphasizes the generation of interpretable biological understanding rather than the reproduction of specific computational outputs.

As illustrated in Figure~\ref{task2}B, the questions in BAIS-SD are organized into several major categories. Some questions correspond to conventional analytical tasks commonly encountered in single-cell studies, such as key gene analysis and cell heterogeneity analysis. 
In parallel, BAIS-SD also includes questions that probe discovery-oriented biological reasoning, such as cellular function reasoning and developmental state analysis.
By jointly incorporating both established analytical tasks and questions derived from published biological discoveries, BAISBench goes beyond evaluating computational proficiency alone and directly assesses whether AI scientists can generate biologically meaningful and interpretable insights from data.

Several example questions are presented below to illustrate the purpose and scope of this task:

\begin{figure*}[t]
  \centering
  \includegraphics[width=0.85\textwidth]{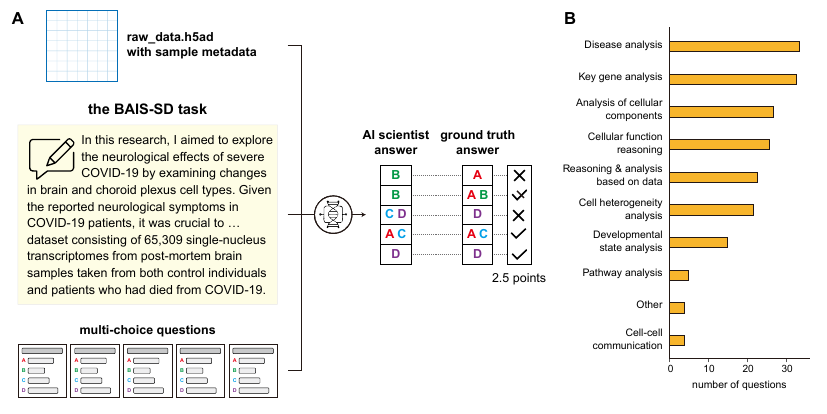}
  \caption{(A) Pipeline of the BAIS-SD task. The AI scientist is provided with background information and a corresponding single-cell dataset, and is required to answer multiple-choice questions by performing data analysis and biological reasoning. Its answers are then compared against the ground truth derived from published literature.
(B) Distribution of question categories in the BAIS-SD multiple-choice question set.}
  \label{task2}
\end{figure*}

\begin{tcolorbox}[
    width=0.475\textwidth,
    arc=2mm, auto outer arc,
    title={\textbf{Examples of multi-choice questions}},breakable,]	
    \textbf{What role did FOXL1+ fibroblasts play in intestinal development, as observed in the data?}
    \begin{itemize}
        \item[(A)] They were involved in crypt-villus differentiation by expressing BMP ligands.
        \item[(B)] They inhibited the differentiation of intestinal stem cells in the fetal gut.
        \item[(C)] They acted as the major source of WNT3A in the developing intestine.
        \item[(D)] They were the primary source of VEGF for endothelial cell development.
    \end{itemize}
    \par\vspace{1em}
    \textbf{Based on a combination of transcriptomic data and external knowledge, which of the following factors are proposed to contribute to increased proliferation in foreskin keratinocytes?}
    \begin{itemize}
        \item[(A)] High expression of amphiregulin (AREG)
        \item[(B)] Increased activation of TGF-${\beta}$ signaling
        \item[(C)] Low levels of CXCL14 and CCL27
        \item[(D)] Presence of CD1C+CD301A+ myeloid dendritic cells
    \end{itemize}

\end{tcolorbox}

In the BAIS-SD task, each AI scientist is provided with a dataset and minimal background context and is required to answer five questions by performing data analysis and biological reasoning (Figure~\ref{task2}A). 
The AI scientists must preprocess the data, conduct relevant analyses using appropriate tools, and reason through the question in order to select the correct options. 
Performance is quantified using an accuracy-based score. For single-choice questions, one point is awarded for a correct answer. For multiple-choice questions, one point is awarded if all correct options are selected, 0.5 points if at least one correct option is selected and no incorrect options are chosen, and zero otherwise. The overall scientific discovery score $S_{\rm SD}$ is defined as:
\begin{equation}
S_{\rm SD} = \frac{n_{\rm single} + n_{\rm multi-all} + 0.5 n_{\rm multi-part}}{n_{\rm Q}}\times 100
\end{equation}
where 
$n_{\rm Q}$ is the total number of questions. 

\subsection{Evaluated AI scientists}

To comprehensively assess the current landscape of AI scientists for biological research and characterize how different system designs affect performance on BAISBench, we evaluated five representative AI scientists spanning a broad range of architectural and methodological choices: AutoBA \citep{zhou2024ai}, scChat \citep{lu2024scchat}, Biomni \citep{huang2025biomni}, Pantheon \citep{pantheon}, and STELLA \citep{jin2025stella}.

\subsubsection{AutoBA}
AutoBA is designed as a lightweight, automation-oriented AI scientist that focuses on constructing end-to-end pipelines for single-cell data preprocessing and analysis. It primarily emphasizes tool orchestration and code generation to execute standard analytical steps with minimal human intervention. 

\subsubsection{scChat}
scChat adopts an interactive, conversation-driven paradigm for single-cell data analysis. By combining a multi-agent architecture with retrieval-augmented generation, scChat enables users to explore datasets through natural language queries while dynamically invoking analytical tools. This design makes scChat particularly suitable for exploratory analysis and human–AI interaction.

\subsubsection{Biomni}
Biomni is a general-purpose AI scientist framework tailored for biomedical research. It integrates literature retrieval, tool-assisted data analysis, and multi-step reasoning within a unified system, allowing it to combine experimental data with external biological knowledge. Biomni is designed to support autonomous biomedical workflows beyond individual tasks, making it more flexible than pipeline-centric systems.

\subsubsection{Pantheon}
Pantheon emphasizes structured scientific reasoning through coordinated multi-agent collaboration. It decomposes complex research objectives into iterative planning, execution, and evaluation steps, enabling systematic exploration of biological hypotheses. Compared to more tool-centric systems, Pantheon places greater emphasis on reasoning consistency and decision-making structure, which can be advantageous for multi-step analytical tasks.

\subsubsection{STELLA}
STELLA represents a hybrid approach that tightly couples LLMs with tool-assisted data analysis and task-specific agent roles. By leveraging multiple base LLM models for different subtasks, STELLA is designed to balance analytical accuracy, reasoning depth, and robustness across diverse biological scenarios. This modular design allows STELLA to adapt to different stages of the biological research workflow, making it particularly well-suited for complex discovery-oriented tasks that require both data analysis and biological interpretation.\\

Together, these systems span a broad design space, ranging from pipeline automation and interactive analysis to structured multi-agent reasoning and hybrid model integration. Evaluating them side by side enables systematic analysis of how architectural choices  influence AI scientists’ ability to support data-driven biological discovery.


\section{Results}

\subsection{The performance of AI scientists on the BAIS-DPTA task}

The BAIS-DPTA task evaluates whether AI scientists can reliably execute standard single-cell analytical workflows and apply biological knowledge to cell type annotation. To enable broader participation and provide an additional reference point, we included a baseline in which GPT-4o directly generated executable Python code for data preprocessing and cell type classification. As a human reference, an graduate-level bioinformatician completed the same task using an automated annotation tool CellTypist \citep{dominguez2022cross} following its official workflow.




\subsubsection{Evaluation of basic data processing and analysis workflows}

As summarized in Table~\ref{tab:task1_performance}, all evaluated AI scientists were able to complete the full preprocessing and analysis pipeline, successfully executing all six required steps: quality control, data normalization, HVG selection, dimensionality reduction, clustering, and marker gene identification. This indicates that current AI scientists have largely mastered the procedural aspects of standard single-cell analysis workflows.

Despite this overall consistency, we observed substantial methodological variation across systems. GPT-4o selected overly stringent quality-control thresholds, resulting in excessive filtering of cells and genes; this issue was manually corrected to ensure fair comparison. 
Only Biomni, Pantheon, and STELLA performed explicit data scaling prior to downstream analysis. scChat adopted the scVI model for dimensionality reduction, whereas other systems relied on principal component analysis (PCA). 
For clustering, Biomni explored multiple resolution parameters, while Pantheon and STELLA uniquely employed the {scipy} implementation of $k$-means clustering; all other systems used the Leiden algorithm implemented in Scanpy . 
Marker gene selection strategies also differed, with most systems selecting three marker genes per cell type, while Biomni selected a broader set of five to eight genes.

Together, these results show that contemporary AI scientists are generally capable of autonomously executing complete and technically sound single-cell preprocessing and analysis pipelines, even though their specific methodological choices vary substantially.

\begin{table*}[!b]
\centering
\caption{Workflow completeness of different AI scientists on the BAIS-DPTA task.
Data norm. denotes data normalization, dim. reduction denotes dimensionality reduction, and \# marker genes denotes the number of marker genes used for cell type annotation.}
\begin{tabular}{
>{\centering\arraybackslash}m{2.5cm}
*{8}{>{\centering\arraybackslash}m{1.4cm}}
}
\hline
 & \makecell{\textbf{Human}\\\textbf{expert}}
 & \makecell{\textbf{Human w/}\\\textbf{CellTypist}}
 & \makecell{\textbf{GPT-4o}}
 & \makecell{\textbf{AutoBA}}
 & \makecell{\textbf{scChat}}
 & \makecell{\textbf{Biomni}}
 & \makecell{\textbf{Pantheon}}
 & \makecell{\textbf{STELLA}} \\
\hline
\textbf{Quality control}      & $\checkmark$ & $\checkmark$ & \ding{53}     & $\checkmark$ & $\checkmark$ & $\checkmark$ & $\checkmark$ & $\checkmark$ \\
\textbf{Data norm.}        & $\checkmark$ & $\checkmark$ & $\checkmark$ & $\checkmark$ & $\checkmark$ & $\checkmark$ & $\checkmark$ & $\checkmark$ \\
\textbf{HVG selection}                  & $\checkmark$ & $\checkmark$ & $\checkmark$ & $\checkmark$ & $\checkmark$ & $\checkmark$ & $\checkmark$ & $\checkmark$ \\
\textbf{Dim. reduction}  & $\checkmark$ & $\checkmark$ & $\checkmark$ & $\checkmark$ & $\checkmark$ & $\checkmark$ & $\checkmark$ & $\checkmark$ \\
\textbf{Clustering}           & $\checkmark$ & --     & $\checkmark$ & $\checkmark$ & $\checkmark$ & $\checkmark$ & $\checkmark$ & $\checkmark$ \\
\textbf{\# marker genes}            & $1\sim3$ & -- & 3 & $2\sim3$ & $1\sim30$ & $5\sim8$ & 3 & $3\sim4$ \\
\hline
\end{tabular}
\label{tab:task1_performance}
\end{table*}

\subsubsection{Cell type annotation performance
}

\begin{figure*}[tb]
  \centering
  \includegraphics[width=1.0\textwidth]{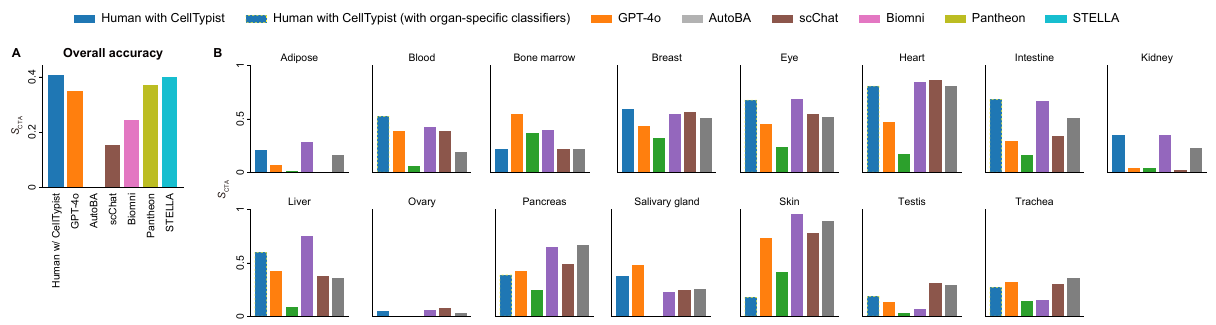}
  \caption{The cell type annotation accuracy of different AI models in the BAIS-DPTA task. (A) Overall results. (B) Results on different organs. The results of AutoBA are not shown in (B) as they are all zero.}
  \label{task1result}
\end{figure*}

In contrast to the uniformly strong performance in workflow execution, we observed greater divergence in cell type annotation accuracy. As shown in Figure~\ref{task1result}A, the human expert using CellTypist achieved the highest overall accuracy. Among the AI scientists, STELLA achieved the best performance, although it remained slightly below the human–CellTypist baseline. AutoBA failed to complete the annotation task, as it was unable to assign cell labels after identifying marker genes.

Notably, several AI scientists, including AutoBA, scChat, and Biomni, performed worse than the baseline approach in which GPT-4o directly generated executable code for cell type classification. This result suggests that increased system complexity or architectural sophistication does not necessarily translate into improved biological interpretation when explicit domain knowledge must be applied.

Analysis across individual organs revealed broadly consistent performance trends (Figure~\ref{task1result}B). Most methods performed well on certain tissues, such as heart, but struggled on others, including ovary. 
Notably, CellTypist provides both organ-specific classifiers trained on dedicated tissue datasets and more general-purpose models trained on broader immune or cross-tissue data. Although organ-specific classifiers are expected to achieve superior annotation accuracy on their corresponding tissues, we found that they did not consistently outperform AI scientists across multiple organs. This observation suggests that AI scientists may possess the capacity to adapt their analysis and incorporate organ-specific knowledge in a data-driven manner, highlighting their potential as flexible alternatives to traditional automated annotation tools.

We further examined the annotation strategies adopted by different systems. Pantheon and STELLA assigned cell types by computing module-level scores using the Scanpy {score\_genes} function and selecting labels based on these scores, whereas other systems primarily relied on overlap between differentially expressed genes and predefined marker genes. These strategy differences were associated with measurable performance differences.

\begin{figure}[htbp]
  \centering
  \includegraphics[width=0.48\textwidth]{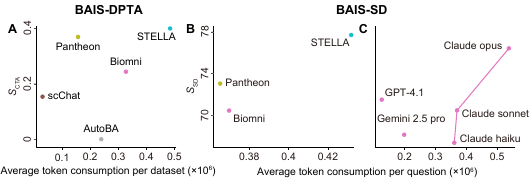}
  \caption{The relationship of AI scientist performance and LLM token consumption in (A) BAIS-DPTA, (B) BAIS-SD using different AI scientists, and (C) BAIS-SD using Biomni with diferent base LLM models.
}
  \label{fig:perfrom_token}
\end{figure}

Finally, we analyzed the relationship between annotation performance and total LLM token consumption (Figure~\ref{fig:perfrom_token}A). Excluding AutoBA, which failed to produce valid annotations, we observed a clear positive correlation between performance and token usage across AI scientists. This trend suggests that, under the current paradigm, improved performance in biologically informed annotation is achieved largely through increased computational and reasoning resources.

\subsubsection{Summary of BAIS-DPTA results
}

Taken together, these results reveal a clear capability gap in current AI scientists. While they are largely proficient at executing standardized single-cell preprocessing and analysis workflows, they remain less reliable than human experts for tasks that require nuanced biological judgment, such as accurate cell type interpretation. At the same time, although CellTypist guided by a human expert achieves the highest overall accuracy, AI scientists outperform it on specific organs, highlighting their potential to complement and, in some cases, surpass human-guided annotation workflows.

\subsection{The performance of  AI scientists on the BAIS-SD task}

The BAIS-SD task evaluates whether AI scientists can generate biologically meaningful insights from real single-cell datasets by answering questions derived from published discoveries. Unlike BAIS-DPTA, this task requires not only correct execution of data analysis pipelines, but also integrative reasoning that combines experimental evidence with biological knowledge.

Not all evaluated AI scientists were able to complete the BAIS-SD task. scChat was excluded because its function tools are hard-coded and do not support the customized analytical workflows required for this benchmark. AutoBA was also excluded, as it consistently failed to produce valid answers. Consequently, our evaluation focused on three AI scientists: Biomni, Pantheon, and STELLA. Biomni and Pantheon both adopted Claude Sonnet 4.5 as their underlying base LLM model, whereas STELLA employed a heterogeneous configuration that combines Claude Sonnet 4.5, Gemini 2.5 Pro, and Grok-4 to handle different subtasks.

To provide a human performance reference, we additionally invited five graduate-level bioinformaticians to independently complete the same question set. Each participant was randomly assigned one-fifth of the questions, collectively covering the full BAIS-SD benchmark.

\subsubsection{Multiple-choice question answering performance}

\begin{figure*}[tb]
  \centering
  \includegraphics[width=0.9\textwidth]{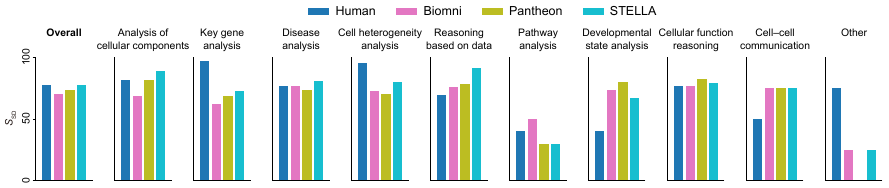}
  \caption{The performance of AI scientists on the BAIS-SD task.}
  \label{task2result}
\end{figure*}

Overall performance on the BAIS-SD task is summarized in Figure~\ref{task2result}. Among the evaluated AI scientists, STELLA achieved the highest overall score, reaching performance comparable to that of human experts. Pantheon ranked second, followed by Biomni. This result indicates that, under appropriate system designs, AI scientists can already approach human-level performance in data-driven biological discovery tasks.

Two factors likely contribute to STELLA’s superior performance. First, its modular design allows different subtasks to be handled by base LLM models best suited to their respective requirements, facilitating more effective integration of data analysis and biological reasoning. Second, STELLA consumed a substantially larger number of tokens than the other systems (Figure~\ref{fig:perfrom_token}B), enabling more exhaustive exploration of the data and reasoning space. Together, these observations suggest that both architectural flexibility and computational budget play important roles in achieving high performance on discovery-oriented tasks.

\subsubsection{Performance across biological question categories
}

To better understand model behavior, we further examined performance across ten question categories (Figure~\ref{task2result}). Despite differences in absolute accuracy, all three AI scientists exhibited similar performance trends, with certain question types consistently easier or more challenging across systems. Notably, STELLA outperformed the other AI scientists in nearly all categories, indicating robust advantages across diverse types of biological reasoning.

Human experts showed clear advantages in key gene analysis and cell heterogeneity analysis, reflecting the importance of domain expertise and interpretative judgment in these tasks. In contrast, AI scientists often matched or exceeded human performance in data-driven reasoning and developmental state analysis, where systematic and exhaustive exploration of high-dimensional data may confer an advantage. These results suggest that AI scientists and human experts exhibit complementary strengths across different aspects of biological discovery.

Taken together, these findings indicate that current AI scientists are already capable of achieving human-comparable performance on several discovery-oriented tasks. Although a gap to the ideal performance ceiling remains, particularly for tasks requiring nuanced biological interpretation, the observed results highlight substantial potential for AI scientists in supporting data-driven biological research.

\subsubsection{Impact of base LLM models on BAIS-SD performance
}

To disentangle the effects of system architecture and underlying base LLM models, we further evaluated the Biomni framework using different LLM models. Specifically, we tested Claude Haiku 3.5 as a weaker baseline, Claude Opus 4.5 as a stronger baseline, and two models from other providers, GPT-4.1 and Gemini 2.5 Pro.

As summarized shown in Figure~\ref{fig:perfrom_token}C, the choice of base LLM model had a pronounced impact on BAIS-SD performance. Within the Claude family, performance followed a clear ordering of Opus $>$ Sonnet $>$ Haiku, with Claude Opus 4.5 elevating Biomni’s performance to a level close to that of human experts. One contributing factor is that stronger models typically operate with higher token budgets, and within the same architecture, AI scientist performance was positively correlated with token usage.

Interestingly, GPT-4.1 and Gemini 2.5 Pro achieved performance comparable to Claude Sonnet 4.5 while requiring fewer tokens (Figure~\ref{fig:perfrom_token}C), suggesting differences in efficiency across base LLM models. Overall, these results indicate that, relative to system architecture, the choice of base LLM models plays a more decisive role in determining performance on BAIS-SD. Stronger base models can directly and substantially enhance AI scientists’ ability to generate biologically meaningful insights from data.

\subsubsection{Summary of BAIS-SD results}

Taken together, the BAIS-SD results demonstrate that current AI scientists are already capable of performing several discovery-oriented tasks at a level comparable to human experts. While human researchers retain clear advantages in tasks that rely heavily on domain intuition and interpretative judgment, such as key gene identification and cell heterogeneity analysis, AI scientists often match or exceed human performance in data-driven reasoning and developmental state analysis. These complementary strengths suggest that AI scientists are particularly effective when systematic exploration of high-dimensional data is required.

At the same time, performance differences across systems and base LLM models indicate that current capabilities remain strongly constrained by the underlying language models and available computational resources. Together, these findings suggest that AI scientists have begun to cross a threshold from purely assistive tools toward meaningful participants in data-driven biological discovery, while still falling short of fully autonomous scientific reasoning.

\section{Discussion}


In this work, we introduce BAISBench as a data-driven benchmark for evaluating AI scientists in realistic biological research scenarios. BAISBench decomposes biological research into two complementary tasks, BAIS-DPTA and BAIS-SD, with the goal of assessing both procedural analytical competence and data-driven biological reasoning. This design reflects our premise that being an AI scientist is not a single capability, but a combination of heterogeneous skills with different levels of maturity. To contextualize system performance, we further include human baselines by involving graduate-level bioinformaticians in the same evaluation tasks.

The results demonstrate that evaluating AI scientists is fundamentally different from evaluating standalone models. AI scientists operate as agentic systems that must plan, execute, and iteratively refine analyses over extended trajectories. As a result, failures often arise from system-level behaviors such as brittle execution, error accumulation, and mismatches between planning and implementation, rather than from isolated reasoning errors. BAISBench is designed to surface these behaviors by grounding evaluation in end-to-end tasks built on real single-cell datasets.


Despite encouraging progress, several recurring capability bottlenecks currently limit the reliability of AI scientists in practice. Many systems can generate plausible high-level plans but struggle to translate them into robust, executable code, particularly in multi-step workflows where small errors can propagate. In addition, reliance on rigid, pre-programmed pipelines constrains generalization to novel datasets or customized analytical goals. Finally, most AI scientists do not proactively integrate external biological knowledge during reasoning, which often leads to shallow interpretation or overly generic conclusions even when the underlying analyses are technically correct. Together, these limitations suggest that current AI scientists are better suited as capable assistants than autonomous agents of discovery.

An important component of BAISBench is the explicit inclusion of human performance baselines. Comparing AI scientists with graduate-level bioinformaticians provides a concrete reference for interpreting benchmark results and grounds system performance in realistic expectations of current research practice. Rather than treating human performance as a fixed upper bound, we view it as a contextual anchor that highlights complementary strengths and weaknesses across different task types. At the same time, we acknowledge that human evaluation in this study remains limited in scale. Due to the substantial time and expertise required for data-driven discovery tasks, questions were partitioned across individuals, and inter-individual variation among human researchers was not systematically assessed. Characterizing such variation represents an important but resource-intensive direction for future work and will likely require broader community involvement.

It is also important to clarify the scope of what BAISBench measures, particularly for the BAIS-SD task. Questions in BAIS-SD are grounded in published discoveries, and the benchmark evaluates whether systems can recover established conclusions from the corresponding datasets. This design does not aim to quantify creativity or novelty directly, and ambiguity or alternative interpretations are inherent to scientific analysis. As such, imperfect scores should not be interpreted as definitive failure, but rather as indicators of alignment with reproducible, evidence-based interpretations reported in the literature.

Taken together, these observations point toward a near-term trajectory in which AI scientists are most effective as part of human–AI collaborative research workflows. AI systems can automate procedural analysis and perform systematic exploration at scale, while human experts contribute biological intuition, contextual interpretation, and selective abstraction. In this context, BAISBench is intended not merely as a leaderboard, but as a diagnostic framework to identify bottlenecks and guide the development of more reliable, generalizable, and biologically grounded AI scientists. We hope BAISBench will help the community track progress, compare emerging systems under realistic conditions, and ultimately enable more effective collaboration between AI and human researchers in data-driven biology.


\bibliographystyle{unsrt}
\bibliography{main}

\clearpage
\pagenumbering{roman}

\renewcommand{\thefigure}{S\arabic{figure}}
\renewcommand{\thetable}{S\arabic{table}}

\setcounter{figure}{0}
\setcounter{table}{0}

\section*{Supplementary Materials}

\subsection{Details of prompts}
\begin{tcolorbox}[
    width=0.475\textwidth,
    arc=2mm, auto outer arc,
    title={\textbf{Prompt for constructing multi-choice questions}},breakable,]	
    {I need to assign an assignment for a bioinformatics analysis class that is about giving a single cell transcriptome data and answering questions by analyzing the given data. I now need to design the questions based on what is in the original article that corresponds to this dataset. I will give you the article below and ask you to read the contents of this article carefully and complete the following tasks:
    \\
    \par
    1. A quick and short summary of the research background in the first person. And the basic information about the sequencing data.
    \\
    \par
    2. Consider which of the conclusions/discoveries in the article are derived directly from the single-cell transcriptome data measured by the authors? Please list them all (use "the data ..." instead of "the study ...", "the author ..." or "the research ..."); this will serve as fodder for my questions. Be as specific as possible, include specific key terms or descriptions, and if necessary, include the process or intermediate steps that led to the conclusion, no generalizations, no descriptive words.
    \\
    \par
    3. Consider which conclusions/discoveries in the article are based on a combination of data measured by the author and external knowledge. Please list them all, this is also my source material for the questions, all requirements are strictly the same as the one above.
    \\
    \par
    4. Choose 5 appropriate conclusions/discoveries from the above and form them into multiple-choice questions (each entry is a separate question). It can be either single (only one correct answer, like "B") or multi-choice (more than one correct answer, like "ACD") questions. Make the position of the correct answer as random as possible. The correct option comes from the article; the incorrect option can come from the article or be added from your own knowledge, but not judged too easily. Give the correct answer to the questions. I would not give the article to the students, so don't come up with anything that need to reading the article That is, avoid expressions like “xxx in the study”, "the author ..." or "xxx in the research", instead, using "in the data ...". What you need to do is treat what is in the article as a standard answer so that students can reproduce those conclusions or discoveries from the given transcriptomic data (not any other data).
    }
    \label{prompt_multichoice}

\end{tcolorbox}


\begin{tcolorbox}[
    width=0.475\textwidth,
    arc=2mm, auto outer arc,
    title={\textbf{Example prompt for AI scientist in BAIS-SD}},breakable,]	
    {I am a bioinformatician and I am doing research on newly measured single-cell transcriptomic data. I now give you the data file (in h5ad format) and the background information about the dataset and my research. I need you to finish the following multiple-choice questions. You need to analyze the data based on the question and give the right option. The answer can be derived from data (some of them may need some extra knowledge, but most are derived mainly from data analysis). The necessary metadata is stored in adata.obs, you can read them and choose the useful ones. Here is the information and the question:
\\
\par
\textbf{Background}:

Lung development is a highly complex process involving a diverse array of cell types, yet our understanding of late-stage human lung development remains incomplete. Animal models have provided critical insights, but translating these findings to human biology is challenging due to species differences. To address this gap, we used single-cell RNA sequencing (scRNA-seq) to create a molecular atlas of newborn human lung cells. This allows us to define distinct cellular populations and their gene signatures, offering new insights into the structural and functional maturation of the human lung at birth.
Sample Source: Two one-day-old newborn human lungs were obtained through organ donation. One was from a full-term infant (38 weeks gestational age), and the other was preterm (31 weeks gestational age).
Cell Isolation \& Processing: Lungs were enzymatically digested to obtain single-cell suspensions, which were frozen and later used for sequencing.
Sequencing Platform: Chromium 10X Genomics system (v2 chemistry), sequenced on a HiSeq4000.
Final Dataset: 5,499 high-quality cells, including epithelial, endothelial, mesenchymal, and immune cells.
\\
\par
\textbf{Question}:

Which major cell type was found to be the most abundant in the newborn human lung based on single-cell transcriptome data?

A) Endothelial cells

B) Epithelial cells

C) Mesenchymal cells

D) Immune cells
\\
\par
Which of the following markers was specifically associated with immature matrix fibroblasts in the newborn lung?

A) SFTPB

B) HES1

C) CDH5

D) PTPRC
\\
\par
Based on single-cell transcriptomic data, what was a key characteristic of immune cells in the newborn human lung?

A) They were only detected in one of the two donors.

B) They were exclusively macrophages.

C) They included T cells, B cells, and macrophages with donor-to-donor variation.

D) They showed no expression of leukocyte markers.
\\
\par
The estimated developmental state of human newborn lung cells, based on murine postnatal development, was closest to which range of murine postnatal days?

A) 1–3 days

B) 4–9 days

C) 10–15 days

D) 16–20 days
\\
\par
What evidence supports the presence of two distinct matrix fibroblast populations in the newborn lung?

A) Differential expression of EPCAM and PECAM1

B) Separation of cells based on mitochondrial gene content

C) Identification of distinct gene expression profiles, including COL6A3 and TCF21

D) Complete absence of mesenchymal markers in one fibroblast population
}
\end{tcolorbox}

\newcolumntype{Y}{>{\raggedright\arraybackslash}X}        
\newcolumntype{Z}{>{\raggedright\arraybackslash}X}        
\newcolumntype{W}{>{\hsize=1.6\hsize\raggedright\arraybackslash}X} 

\begin{table*}[!h]
\centering
\footnotesize
\caption{Summary of single-cell datasets used in the BAIS-DPTA task.}
\begin{tabularx}{\textwidth}{
Y                          
>{\centering\arraybackslash}p{0.7cm}       
Y                          
Y                          
W                          
r                          
r                          
}
\toprule
\textbf{Paper} & \textbf{Year} & \textbf{Journal} & \textbf{Organ} & \textbf{DOI} &
\makecell{\textbf{Cell} \\ \textbf{number}} &
\makecell{\textbf{Cell type} \\ \textbf{number}} \\
\midrule
Aizarani et al. & 2019 & \textit{Nature} & Liver &
\href{https://doi.org/10.1038/s41586-019-1373-2}{10.1038/s41586-019-1373-2} &
9194 & 12 \\

He et al. & 2020 & \textit{Genome Biology} & Intestine &
\href{https://doi.org/10.1186/s13059-020-02210-0}{10.1186/s13059-020-02210-0} &
8924 & 12 \\

Chitiashvili et al. & 2020 & \textit{Nature Cell Biology} & Ovary &
\href{https://doi.org/10.1038/s41556-020-00607-4}{10.1038/s41556-020-00607-4} &
8561 & 6 \\

Cao et al. & 2020 & \textit{Science} & Pancreas &
\href{https://doi.org/10.1126/science.aba7721}{10.1126/science.aba7721} &
43155 & 14 \\

Zhao et al. & 2020 & \textit{Nature Commun} & Testis &
\href{https://doi.org/10.1038/s41467-020-19414-4}{10.1038/s41467-020-19414-4} &
26482 & 8 \\

Miller et al. & 2020 & \textit{Developmental Cell} & Trachea &
\href{https://doi.org/10.1016/j.devcel.2020.01.033}{10.1016/j.devcel.2020.01.033} &
17423 & 5 \\

Roy et al. & 2021 & \textit{Cell Reports} & Bone marrow &
\href{https://doi.org/10.1016/j.celrep.2021.109698}{10.1016/j.celrep.2021.109698} &
30894 & 6 \\

Voigt et al. & 2021 & \textit{Hum Mol Genet} & Eye &
\href{https://doi.org/10.1093/hmg/ddab140}{10.1093/hmg/ddab140} &
31870 & 10 \\

Emont et al. & 2022 & \textit{Nature} & Adipose &
\href{https://doi.org/10.1038/s41586-022-04518-2}{10.1038/s41586-022-04518-2} &
55150 & 10 \\

Domínguez Conde et al. & 2022 & \textit{Science} & Blood &
\href{https://doi.org/10.1126/science.abl5197}{10.1126/science.abl5197} &
24149 & 26 \\

Suo et al. & 2022 & \textit{Science} & Kidney &
\href{https://doi.org/10.1126/science.abo0510}{10.1126/science.abo0510} &
25955 & 42 \\

Tabula Sapiens Consortium et al. & 2022 & \textit{Science} & Breast &
\href{https://doi.org/10.1126/science.abl4896}{10.1126/science.abl4896} &
11227 & 13 \\

Tabula Sapiens Consortium et al. & 2022 & \textit{Science} & Salivary gland &
\href{https://doi.org/10.1126/science.abl4896}{10.1126/science.abl4896} &
26959 & 22 \\

Gur et al. & 2022 & \textit{Cell} & Skin &
\href{https://doi.org/10.1016/j.cell.2022.03.011}{10.1016/j.cell.2022.03.011} &
38666 & 8 \\
\bottomrule
\end{tabularx}
\label{task1data}
\end{table*}



\begin{table*}[!p]
\centering
\footnotesize
\renewcommand{\arraystretch}{1.2}
\caption{Summary of single-cell papers and datasets used in the Scientific discovery task.}
\begin{tabularx}{\textwidth}{
>{\raggedright\arraybackslash}X      
>{\centering\arraybackslash}p{0.7cm} 
>{\raggedright\arraybackslash}X      
>{\raggedright\arraybackslash}X      
r                                     
}
\toprule
\textbf{Paper} & \textbf{Year} & \textbf{Journal} & \textbf{DOI} & \textbf{Cell number} \\
\midrule
Fan et al. & 2019 & \textit{Nat Commun} &
\href{https://doi.org/10.1038/s41467-019-11036-9}{10.1038/s41467-019-11036-9} &
20676 \\

J\"akel et al. & 2019 & \textit{Nature} &
\href{https://doi.org/10.1038/s41586-019-0903-2}{10.1038/s41586-019-0903-2} &
17799 \\

Martin et al. & 2019 & \textit{Cell} &
\href{https://doi.org/10.1016/j.cell.2019.08.008}{10.1016/j.cell.2019.08.008} &
32458 \\

Menon et al. & 2019 & \textit{Nat Commun} &
\href{https://doi.org/10.1038/s41467-019-12780-8}{10.1038/s41467-019-12780-8} &
20091 \\

Stewart et al. & 2019 & \textit{Science} &
\href{https://doi.org/10.1126/science.aat5031}{10.1126/science.aat5031} &
105870 \\

Szabo et al. & 2019 & \textit{Nat Commun} &
\href{https://doi.org/10.1038/s41467-019-12464-3}{10.1038/s41467-019-12464-3} &
51876 \\

Cowan et al. & 2020 & \textit{Cell} &
\href{https://doi.org/10.1016/j.cell.2020.08.013}{10.1016/j.cell.2020.08.013} &
98348 \\

Elmentaite et al. & 2020 & \textit{Developmental Cell} &
\href{https://doi.org/10.1016/j.devcel.2020.11.010}{10.1016/j.devcel.2020.11.010} &
85351 \\

Joseph et al. & 2020 & \textit{Prostate} &
\href{https://doi.org/10.1002/pros.24020}{10.1002/pros.24020} &
122129 \\

Lavaert et al. & 2020 & \textit{Immunity} &
\href{https://doi.org/10.1016/j.immuni.2020.03.019}{10.1016/j.immuni.2020.03.019} &
71732 \\

Lee et al. & 2020 & \textit{Sci Immunol} &
\href{https://doi.org/10.1126/sciimmunol.abd1554}{10.1126/sciimmunol.abd1554} &
59572 \\

Lukassen et al. & 2020 & \textit{The EMBO Journal} &
\href{https://doi.org/10.15252/embj.20105114}{10.15252/embj.20105114} &
57229 \\

Sol\'e-Boldo et al. & 2020 & \textit{Commun Biol} &
\href{https://doi.org/10.1038/s42003-020-0922-4}{10.1038/s42003-020-0922-4} &
15457 \\

Wang et al. & 2020 & \textit{J Exp Med} &
\href{https://doi.org/10.1084/jem.20191130}{10.1084/jem.20191130} &
14106 \\

Wu et al. & 2020 & \textit{The EMBO Journal} &
\href{https://doi.org/10.15252/embj.2019104063}{10.15252/embj.2019104063} &
24271 \\

Xiang et al. & 2020 & \textit{Front Cardiovasc Med} &
\href{https://doi.org/10.3389/fcvm.2020.00052}{10.3389/fcvm.2020.00052} &
9980 \\

Melms et al. & 2021 & \textit{Nature} &
\href{https://doi.org/10.1038/s41586-021-03569-1}{10.1038/s41586-021-03569-1} &
116313 \\

Yang et al. & 2021 & \textit{Nature} &
\href{https://doi.org/10.1038/s41586-021-03710-0}{10.1038/s41586-021-03710-0} &
65309 \\

Burclaff et al. & 2022 & \textit{Cell Mol Gastroenterol Hepatol} &
\href{https://doi.org/10.1016/j.jcmgh.2022.02.007}{10.1016/j.jcmgh.2022.02.007} &
12590 \\

Fasolino et al. & 2022 & \textit{Nat Metab} &
\href{https://doi.org/10.1038/s42255-022-00531-x}{10.1038/s42255-022-00531-x} &
69645 \\

Knight-Schrijver et al. & 2022 & \textit{Nat Cardiovasc Res} &
\href{https://doi.org/10.1038/s44161-022-00183-w}{10.1038/s44161-022-00183-w} &
60668 \\

Lengyel et al. & 2022 & \textit{Cell Reports} &
\href{https://doi.org/10.1016/j.celrep.2022.111838}{10.1016/j.celrep.2022.111838} &
86708 \\

Opasawatchai et al. & 2022 & \textit{Front Dent Med} &
\href{https://doi.org/10.3389/fdmed.2021.806294}{10.3389/fdmed.2021.806294} &
6560 \\

Watanabe et al. & 2022 & \textit{Am J Respir Cell Mol Biol} &
\href{https://doi.org/10.1165/rcmb.2021-0555OC}{10.1165/rcmb.2021-0555OC} &
57918 \\

Xu et al. & 2022 & \textit{Sci Rep} &
\href{https://doi.org/10.1038/s41598-022-17832-6}{10.1038/s41598-022-17832-6} &
36908 \\

Horeth et al. & 2023 & \textit{J Dent Res} &
\href{https://doi.org/10.1177/00220345221147908}{10.1177/00220345221147908} &
15684 \\

Kurkalang et al. & 2023 & \textit{Cancer Science} &
\href{https://doi.org/10.1111/cas.15979}{10.1111/cas.15979} &
28186 \\

Rustam et al. & 2023 & \textit{Am J Respir Crit Care Med} &
\href{https://doi.org/10.1164/rccm.202207-1384OC}{10.1164/rccm.202207-1384OC} &
115788 \\

Strati et al. & 2023 & \textit{Cell Reports Medicine} &
\href{https://doi.org/10.1016/j.xcrm.2023.101158}{10.1016/j.xcrm.2023.101158} &
92676 \\

Whitfield et al. & 2023 & \textit{Clinical \& Translational Med} &
\href{https://doi.org/10.1002/ctm2.1356}{10.1002/ctm2.1356} &
62599 \\

Wiedemann et al. & 2023 & \textit{Cell Reports} &
\href{https://doi.org/10.1016/j.celrep.2023.111994}{10.1016/j.celrep.2023.111994} &
15243 \\

Bhattacharya et al. & 2024 & \textit{Genes} &
\href{https://doi.org/10.3390/genes15030298}{10.3390/genes15030298} &
5499 \\

Binvignat et al. & 2024 & \textit{JCI Insight} &
\href{https://doi.org/10.1172/jci.insight.178499}{10.1172/jci.insight.178499} &
108717 \\

de Vrij et al. & 2024 & \textit{Commun Biol} &
\href{https://doi.org/10.1038/s42003-024-06225-2}{10.1038/s42003-024-06225-2} &
30130 \\

Guerrero-Murillo et al. & 2024 & \textit{bioRxiv} &
\href{https://doi.org/10.1101/2024.01.23.576878}{10.1101/2024.01.23.576878} &
37100 \\

Heimlich et al. & 2024 & \textit{Blood Advances} &
\href{https://doi.org/10.1182/bloodadvances.2023011445}{10.1182/bloodadvances.2023011445} &
66985 \\

Li et al. & 2024 & \textit{Cell Stem Cell} &
\href{https://doi.org/10.1016/j.stem.2023.12.013}{10.1016/j.stem.2023.12.013} &
67996 \\

Mimpen et al. & 2024 & \textit{The FASEB Journal} &
\href{https://doi.org/10.1096/fj.202300601RRR}{10.1096/fj.202300601RRR} &
10533 \\

Moerkens et al. & 2024 & \textit{Cell Reports} &
\href{https://doi.org/10.1016/j.celrep.2024.114247}{10.1016/j.celrep.2024.114247} &
22280 \\

Phan et al. & 2024 & \textit{Nat Commun} &
\href{https://doi.org/10.1038/s41467-024-45165-7}{10.1038/s41467-024-45165-7} &
98848 \\

Rabadam et al. & 2024 & \textit{JCI Insight} &
\href{https://doi.org/10.1172/jci.insight.176963}{10.1172/jci.insight.176963} &
105827 \\
\bottomrule
\end{tabularx}
\label{task2data}
\end{table*}

\end{document}